\documentclass[letterpaper, 10 pt, conference]{ieeeconf}  

\IEEEoverridecommandlockouts                              
\usepackage{graphicx}      
\usepackage{algorithm} 
\usepackage{algpseudocode} 
\usepackage{varwidth} 
\usepackage{cite}
\usepackage{amsfonts}
\usepackage{amsmath}
\usepackage{url}
\usepackage{gensymb}
\usepackage[dvipsnames]{xcolor}

\usepackage[caption=false,font=footnotesize]{subfig}

\graphicspath{{figs/}}
\title{\LARGE \bf
Learn to Grasp with Less Supervision: \\ A Data-Efficient Maximum Likelihood Grasp Sampling Loss
}

\author{Xinghao Zhu$^1$, Yefan Zhou$^1$, Yongxiang Fan$^2$, Lingfeng Sun$^1$, Jianyu Chen$^1$, and Masayoshi Tomizuka$^1$
\thanks{$^1$Mechanical Systems Control Lab, UC Berkeley, CA USA. 
{\tt\small \{zhuxh, yefan0726, lingfengsun, jianyuchen, tomizuka\}@berkeley.edu}}
\thanks{$^2$FANUC Advanced Research Lab, FANUC America Corporation, CA USA.
{\tt\small Yongxiang.Fan@fanucamerica.com}}
\thanks{$^*$Labels refer to marking the image to indicate a successful robotic grasp.}
}

\begin{document}
\maketitle
\thispagestyle{empty}
\pagestyle{empty}

\begin{abstract}
Robotic grasping for a diverse set of objects is essential in many robot manipulation tasks.
One promising approach is to learn deep grasping models from large training datasets of object images and grasp labels.
However, empirical grasping datasets are typically sparsely labeled (i.e., a small number of successful grasp labels$^*$ in each image).
The data sparsity issue can lead to insufficient supervision and false-negative labels, and thus results in poor learning results.
This paper proposes a Maximum Likelihood Grasp Sampling Loss (MLGSL) to tackle the data sparsity issue.
The proposed method supposes that successful grasps are stochastically sampled from the predicted grasp distribution and maximizes the observing likelihood.
MLGSL is utilized for training a fully convolutional network that generates thousands of grasps simultaneously.
Training results suggest that models based on MLGSL can learn to grasp with datasets composing of 2 labels per image. 
Compared to previous works, which require training datasets of 16 labels per image, MLGSL is 8$\times$ more data-efficient.
Meanwhile, physical robot experiments demonstrate an equivalent performance at a 90.7\% grasp success rate on household objects.
Codes and videos are available at~\cite{website}.
\end{abstract}


\section{Introduction}
\label{sec: introduction}

Robotic grasping in unstructured environments can benefit applications from warehouse automation to home servicing.
Supervised machine learning approaches have demonstrated promising results in planning grasps under various uncertainties.
One kind of approach is to sample grasp candidates and evaluate~\cite{dexnet2, dexnet3, 6dofgraspnet, 6dofgraspnet_inclutter, gpd, ppojpo, finger_splitting, surface_fitting, pointgpd, 2019Lu}.
Such two-step methods, however, might be time-consuming at execution.


An alternative is to train grasp planning models in end-to-end manners~\cite{unigrasp,cgpn,pointnetgrasp, ggcnn, ggcnn2, andy_zeng_grasp,vgn, fcgqcnn, 2015Redmon, 2015Varley}.
These approaches directly generate grasps and have a shorter planning time.
While end-to-end models typically require densely labeled ground-truth samples in training, existing empirical datasets, generated by physical execution or human labeling, only include scarce labels for each image~\cite{jacquard, cornell_dataset}.
Though many other good grasps exist, they are unfortunately not labeled.
The sparsity issue can lead to a lot of false negatives in training end-to-end models.
Previous works~\cite{pointnetgrasp,ggcnn, ggcnn2,andy_zeng_grasp,vgn} assume that unlabeled points are not valid and treat them as failed grasps.
However, robust grasps may still exist in these areas but are mistakenly labeled as negatives.
These wrong labels can harm the learning results.

\begin{figure}[tb]
\begin{center}
	\includegraphics[width=3.4in]{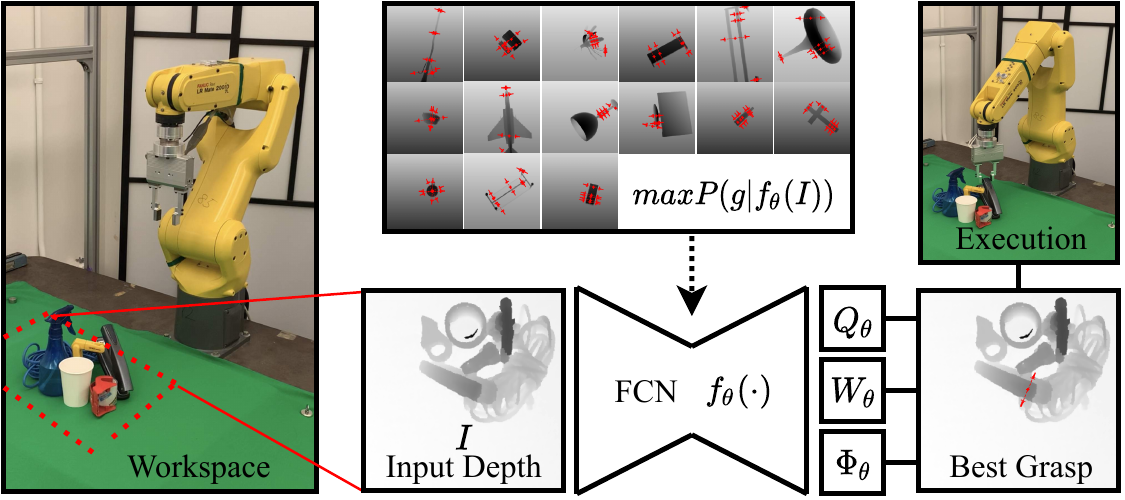}
	\vspace{-1em}
	\caption{Grasp planning and execution pipeline.
	When an object is presented in the workspace, a stereo camera captures a depth image; a trained generative model $f_\theta(\cdot)$ rapidly computes grasp configuration maps $Q_\theta$, $W_\theta$, and $\Phi_\theta$. The best grasp is generated based on configuration maps and executed with the robot manipulator. 
	The grasp model is trained offline with empirical datasets and the proposed loss function.}
	\label{fig: framework}
\end{center}
\end{figure}

This paper proposes a maximum likelihood grasp sampling loss (MLGSL) to improve data efficiency in training grasp planners with single-view depth images.
The main difference between MLGSL and other works is the assumption toward unlabeled regions.
Previous works~\cite{pointnetgrasp,ggcnn, ggcnn2,andy_zeng_grasp,vgn} regard unlabeled areas as failures and estimate the success rate for each pixel.
In contrast, we leave unlabeled pixels intact and solely apply supervisions with labeled grasps.
Specifically, we propose a stochastic grasp selection process to estimate the likelihood for each pixel to be the best grasp point and maximize such likelihood for labeled grasps.
Since no labels are generated for unlabeled areas, MLGSL can reduce the false-negative problem in training grasp planners.
Training results demonstrate that models with MLGSL can learn to grasp with fewer labels compared to previous works~\cite{ggcnn,ggcnn2,andy_zeng_grasp,andy_grasp_push}, while physical experiments show a similar grasp success rate at 90.7\% (Fig.~\ref{fig: framework}).
Moreover, this paper demonstrates that attention mechanisms do not contribute to dense grasp plannings.
Furthermore, a dataset is constructed with multiple-object scenes and collision-free grasp labels to improve the performance in clutter.

Related works are introduced in Section~\ref{sec: related_works}.
Section~\ref{sec: algorithms} presents the proposed approach.
Training and experiments are presented in Section~\ref{sec: training} and~\ref{sec: exp}.
Section~\ref{sec: conclusion} concludes the paper and suggests the future work.




\section{Related Works}
\label{sec: related_works}
\subsection{Grasp Planning and Datasets}
Analytic and machine learning methods have been studied to plan grasps across various objects. 
Existing analytic methods~\cite{quality_metrics, template_matching, 206book} can be used to search for optimal grasps. These techniques, however, are less robust in practice due to perceptional limitations and unseen geometries.

Machine learning is an alternative approach to plan grasps.
Current methods show that it is preferable to learn grasp quality functions and optimize them at the runtime~\cite{dexnet2, 6dofgraspnet, 6dofgraspnet_inclutter, gpd, ppojpo, finger_splitting, surface_fitting, pointgpd,2019Lu}.
However, sampling or optimization makes the algorithm time-consuming and requires pre-defined heuristics.
Other end-to-end approaches propose to infer grasp poses from the raw input directly.
These approaches can be generative or discriminative.
Generative models~\cite{pointnetgrasp, unigrasp, ggcnn, ggcnn2, cgpn, roi_grasp, rpn_grasp_2, rpn_grasp_4, rpn_grasp} perform grasp pose regression and grasp quality assessment simultaneously.
Discriminative approaches~\cite{fcgqcnn, andy_zeng_grasp} use fully convolutional networks to evaluate thousands of grasps simultaneously without direct sampling.
Reinforcement learning has also been introduced in grasp planning~\cite{andy_grasp_push, Q2Opt}. 
Q learning is leveraged to estimate the state-action qualities with iterative updates.

Machine learning approaches typically require large datasets consisting of sensor readings and ground-truth grasp labels. 
Synthetic datasets can be rapidly generated using analytic quality metrics and simulated sensors~\cite{dexnet2, dexnet3, cgpn}. 
However, synthetic data can cause robustness issues due to the simulation-to-reality (sim-to-real) gap. Specifically, analytic metrics might disagree with physical grasp results in complex circumstances.
Alternatively, empirical methods collect data from human labeling or grasp execution with correlations of physical success~\cite{6dofgraspnet, 6dofgraspnet_inclutter, jacquard, cornell_dataset}. Nevertheless, empirical data can be expensive to acquire and are typically sparsely labeled (i.e., inadequate grasp labels in each image). 
This paper chooses to use empirical datasets to avoid the sim-to-real gap.

\subsection{Dense Grasp Planning}
\label{subsec: dense_plan}
Recent works leverage dense pixel-wise evaluation and regression for grasp planning. 
These approaches utilize Fully Convolutional Networks (FCNs) to evaluate millions of grasps in parallel.
FCNs in~\cite{ggcnn, ggcnn2} predict the grasp success rate and generate grasp configurations for each pixel.
In~\cite{andy_zeng_grasp, fcgqcnn}, FCNs are trained to predict the grasp success rate for pre-defined grasp primitive actions.

During training, FCNs require sufficient labels to perform the image-to-image learning~\cite{andy_zeng_grasp}.
However, existing datasets typically include scarce grasp labels for each image~\cite{jacquard, cornell_dataset}.
Some works have been proposed to resolve the sparsity issue.
A label generation method is proposed in~\cite{ggcnn,ggcnn2}, which assumes that grasps close to successful labels are robust and treats unlabeled grasps as failures.
However, this approach introduces both false negatives and false positives.
On the one hand, robust grasps might be away from existing labels and are wrongly labeled as negatives.
On the other hand, generated grasps that are close to existing successful labels might be unstable.
In~\cite{fcgqcnn, 2015Varley}, networks are converted from discriminative evaluators according to the injective mapping between convolutional and fully connected layers. 
A grasp-push policy represented by FCNs is trained in~\cite{andy_grasp_push} with Q learnings. 
Although these methods eliminate the need for dense labels, they still require large datasets or rollouts.
This paper proposes a loss function capable of training the FCN from scratch with inadequate labels.

\subsection{Attention Mechanism}
Attention mechanisms have achieved promising results in computer vision~\cite{caption, sam, detection}.
Previous works have introduced spatial attention mechanism (SAM) to robotics for interest-region extractions and feature reductions.
In~\cite{hsa, hsa2}, hierarchical SAM is employed to constrain sampling in reinforcement learning (RL) and demonstrated improved efficiency in action-space sampling. 
SAM is used in~\cite{sam_grasp, sam_grasp2} to detect grasps in clutter, achieving improved grasping success rate with RL. 
However, experiments in this paper demonstrate that attention mechanisms do not contribute to dense grasp planning with FCNs.

\section{Grasp Planning with Maximum Likelihood Grasp Sampling Loss}
\label{sec: algorithms}
This section first introduces notations of the problem.
The proposed loss function is then illustrated and compared with previous works.
Finally, the network architectures and datasets are presented.

\subsection{Notations}
\label{subsec: notations}
The grasp planning problem is defined as detecting a grasp that allows the robot to pick up objects.
Moreover, no explicit knowledge of the object is given beyond camera readings.

\begin{figure}[tb]
\begin{center}
	\includegraphics[width=1.9in]{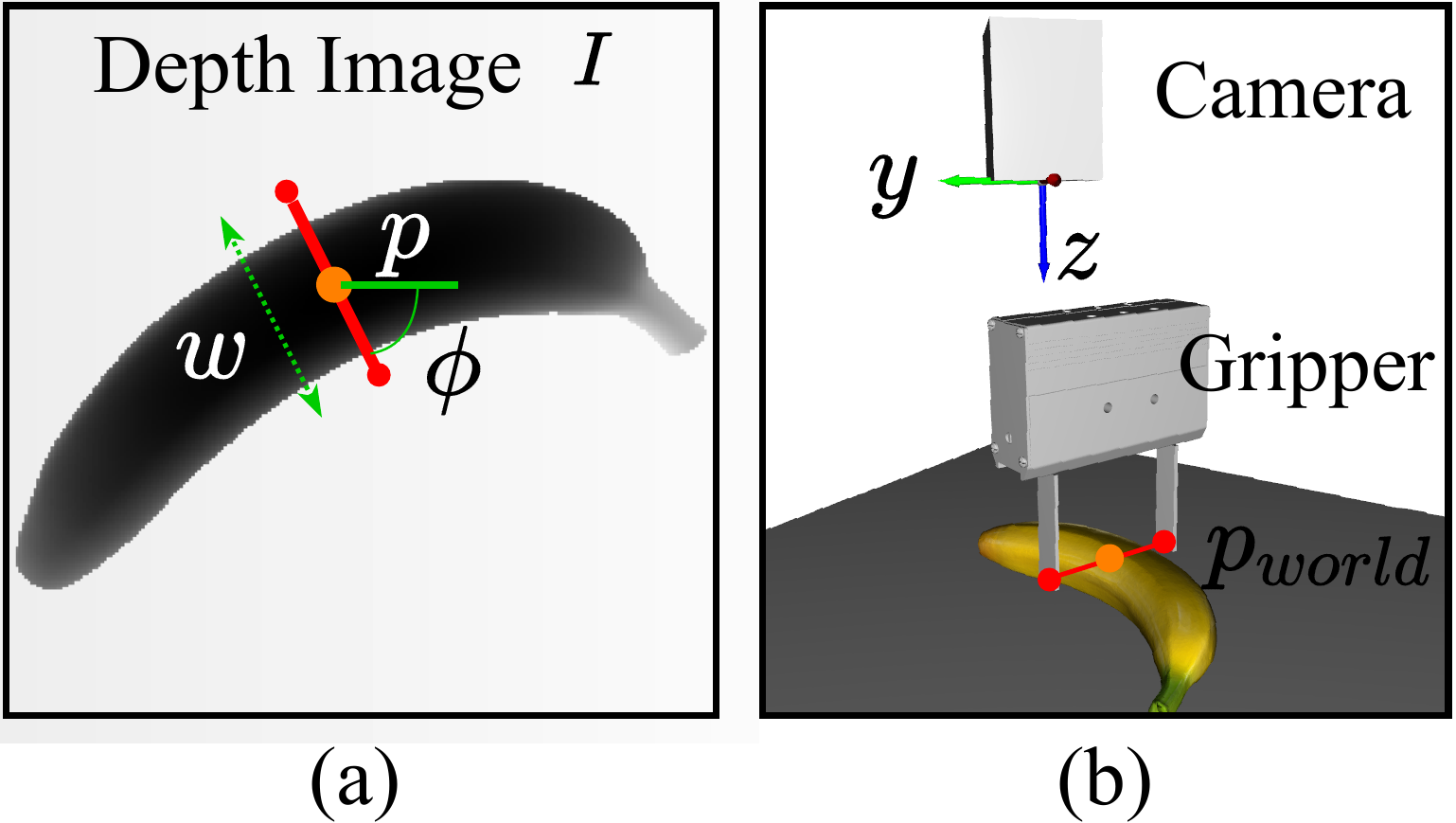}
	\vspace{-0.75em}
	\caption{Grasp Representation $g=(p,\phi, w)$. 
	The planar pose $(p, \phi, w)$ in (a) represents the grasp's center position, orientation, and width in the image. 
	Grasp $g$ in (a) has quality $q=1$ since it is a successful grasp.
	$g$ is executed perpendicular to the image plane at point $p_{world}$ in the Cartesian frame as shown in (b), where $p_{world}$ is $p$ in the world frame.
	The gripper further moves $\epsilon$ cm below $p_{world}$ in the direction of the camera's z-axis.}
	\label{fig: grasp_representation}
\end{center}
\end{figure}

\subsubsection{Grasp}
Let $I \in \mathbb{R}^{\mathcal{H} \times \mathcal{W}}$ define a given depth image with height $\mathcal{H}$ and width $\mathcal{W}$.
The $i$-th grasp is defined in the image $I$ and denoted by 
$g_i=(p_i,\phi_i,w_i)$, where $p_i$ is a pixel in the image representing the grasp center. $\phi_i \in [-\frac{\pi}{2}, \frac{\pi}{2}]$ is the gripper's rotation, and $w_i \in [0, 150]$ pixels is the gripper's width in the image (Fig.~\ref{fig: grasp_representation}). 
Each grasp has a quality measurement $q_i \in [0, 1]$ that indicates the grasp success rate. 

\subsubsection{Grasp Labels}
Empirical datasets contain $k$ success grasp labels $\tilde{g}$ for each image, where $\tilde{g}=\{\tilde{g_i}\}$ for $i \in [1, ..., k]$ and $k \ll \mathcal{H} \times \mathcal{W}$.
Note $\tilde{q_{i}}=1$ since they are guaranteed to succeed.

\subsubsection{Grasp Configuration Maps}
Similar to~\cite{ggcnn}, we refer to the set of grasps in the image as the grasp configuration map $G = (Q, \Phi, W) \in \mathbb{R}^{3 \times \mathcal{H} \times \mathcal{W}}$, where $Q, \Phi, W \in \mathbb{R}^{\mathcal{H} \times \mathcal{W}}$ contain values of $q_i, \phi_i, w_i$ at each pixel of $I$.
In practice, we use two components $\Phi_s = \sin(2\Phi), \Phi_c = \cos(2\Phi)$ for $\Phi$ to resolve the symmetricity of antipodal grasps.

\begin{figure}[tb]
\begin{center}
	\includegraphics[width=3.4in]{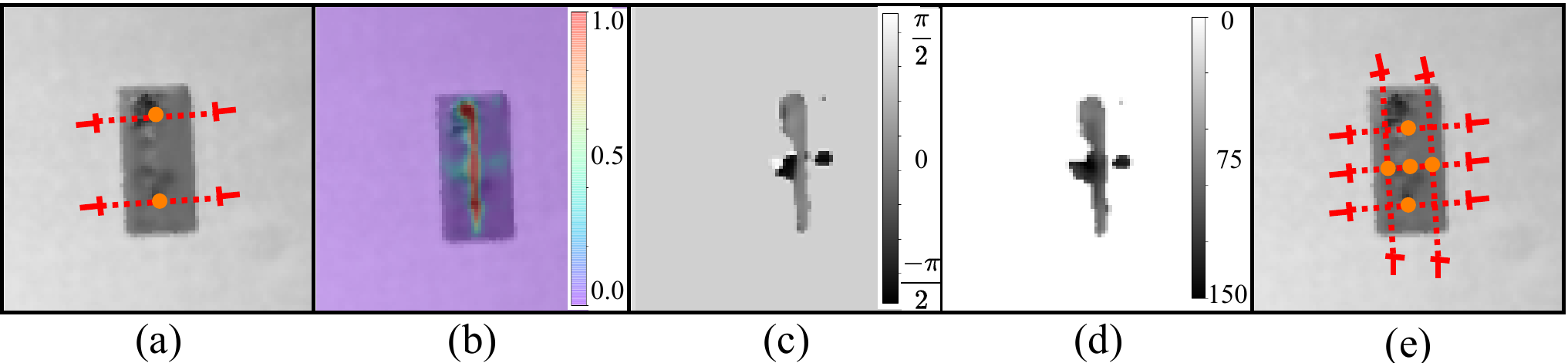}
	\vspace{-1.5em}
	\caption{
	(a) shows the input depth image $I$ with $k=2$ success grasp labels $\tilde{g}$ (red).
	(b-d) show predicted grasp configuration maps $\hat{G}=(\hat{Q}, \hat{\Phi}, \hat{W})$ respectively.
	Different colors represent different values as in the color bar.
	(e) shows 5 selected grasps based on grasp configuration maps $\hat{G}$.
	}
	\label{fig: dataset_sample}
\end{center}
\end{figure}

\subsubsection{Grasp Planning Models}
This paper uses a grasp neural network $f_\theta(\cdot)$ to approximate the dense grasp configuration maps, $\hat{G}=(\hat{Q}, \hat{\Phi}, \hat{W})=f_\theta(I)$.
Predicted grasps are $\hat{g} = \{ \hat{g_i}\}$ for $i \in [1, ..., \mathcal{H}\times \mathcal{W}]$ at each pixel.


\subsection{Maximum Likelihood Grasp Sampling Loss}
\label{subsec: loss}
Following definitions in Section~\ref{subsec: notations}, this paper uses two grasp notations for each image $I$: the predicted grasp configuration maps $\hat{G}$ and the grasp labels $\tilde{g}$.
The former maps are densely predicted by the network, which have values at each pixel.
In contrast, the latter $\tilde{g}$ only has $k$ successful grasp
labels, as shown in Fig.~\ref{fig: dataset_sample}.
The objective of the grasp model is to estimate grasp configuration maps by $\hat{G}=f_\theta(I)$, which yield the highest grasp success rate.

At execution, grasps are selected based on the predictions $\hat{G}$.
First, a grasp pixel $p$ is drawn from the image conditioned on the quality map $\hat{Q}$.
Categorical distribution is used to describe the sampling process. 
Each pixel is a category with a probability proportional to the predicted quality $\hat{q_i}$.
This suggests that for each grasp pixel, the higher the grasp quality is, the more chance it can be selected to execute.
Then, corresponding rotation $\hat{\phi}$ and width $\hat{w}$ are selected from dense prediction maps $\hat{\Phi}$ and $\hat{W}$ at pixel $p$.


This paper assumes that success grasp labels $\tilde{g}$ are chosen from the prediction maps $\hat{G}$ based on the above selection procedure.
To optimize the learning performance, we propose to maximize the chance of observing $\tilde{g}$ from $\hat{G}$ using a maximum likelihood estimation (MLE).

The grasp selection process is first modeled mathematically.
The grasp configuration $\boldsymbol{p}, \boldsymbol{\phi}, \boldsymbol{w}$ are regarded as random variables.
The pixel location $\boldsymbol{p}$ is discrete and takes on a value in the pixel indexes, i.e., $p\in [1, ..., \mathcal{H} \times \mathcal{W}]$.
The probability mass function of $\boldsymbol{p}$ is a Categorical distribution defined as $P(p | \hat{Q})$, representing the chance that the grasp at pixel $p$ is selected for execution.
The rotation $\boldsymbol{\phi}$ and the width $\boldsymbol{w}$ are real values in the configuration space, whose density functions are Gaussians as $P(\phi | p, \hat{\Phi})$ and $P(w | p, \hat{W})$.

Based on the above model, the probability of observing a grasp label $\tilde{g_{i}}$ from the prediction map $\hat{G}$ is $P(\tilde{g_{i}} | \hat{G})$. The weights of the network $\theta$ are trained to maximize such a probability:
\begin{subequations} 
\begin{align}
    & \max_\theta && P(\tilde{g_i} | \hat{G}) \nonumber \\
  = & \max_\theta && P(\tilde{p_i}, \tilde{\phi_i}, \tilde{w_i} | \hat{G}) \label{eqn: decompose_single_objective_1}\\
  = & \max_\theta && P(\tilde{p_i} | \hat{G}) \cdot P(\tilde{\phi_i} |\tilde{p_i}, \hat{G}) \cdot P(\tilde{w_i} |\tilde{\phi_i}, \tilde{p_i}, \hat{G}) \label{eqn: decompose_single_objective_2} \\
  = & \max_\theta && P(\tilde{p_i} | \hat{Q}) \cdot P(\tilde{\phi_i} |\tilde{p_i}, \hat{\Phi}) \cdot P(\tilde{w_i} | \tilde{p_i}, \hat{W}) \label{eqn: decompose_single_objective_3}
\end{align}
\end{subequations}

In~(\ref{eqn: decompose_single_objective_1}), $\tilde{g_i}$ is replaced with grasp configurations as in~\ref{subsec: notations}.
Chain rules are applied to obtain~(\ref{eqn: decompose_single_objective_2}).
Equation~(\ref{eqn: decompose_single_objective_3}) is modified based on the grasp selection model.
The first term maximizes the chance to select a successful grasp at pixel $\tilde{p_i}$ for execution.
The second and third terms maximize the chance of observing $\tilde{\phi_i}$ and $\tilde{w_i}$ at pixel $\tilde{p_i}$ in $\hat{\Phi}$ and $\hat{W}$.

Equation (\ref{eqn: decompose_single_objective_3}) is then jointly optimized for all grasp labels $\tilde{g}$.
This paper further assumes that each label is independent:
\begin{subequations} 
\begin{align}
    & \max_\theta && \prod_{i=1}^k P(\tilde{g_i} | \hat{G}) \\
  \approx & \max_\theta && \prod_{i=1}^k P(\tilde{p_i} | \hat{Q}) \cdot P(\tilde{\phi_i} |\tilde{p_i}, \hat{\Phi}) \cdot P(\tilde{w_i} | \tilde{p_i}, \hat{W} ) \label{eqn: final_objective_2} \\ 
  \propto & \max_\theta && \sum_{i=1}^k \log P(\tilde{p_i} | \hat{Q}) + \log P(\tilde{\phi_i} |\tilde{p_i}, \hat{\Phi})  \label{eqn: final_objective_3} \\ 
  & && \ \ \ \ \  + \log P(\tilde{w_i} | \tilde{p_i}, \hat{W} )  \nonumber \\
  = & \max_\theta && \sum_{i=1}^k \log P(\tilde{p_i} | \hat{Q}) - \text{MSE}(\hat{\phi_i}, \tilde{\phi_i}) \label{eqn: final_objective_4} \\ 
  & && \ \ \ \ \  - \text{MSE}(\hat{w_i}, \tilde{w_i}) \nonumber
\end{align} 
\end{subequations}

Three terms are maximized in~(\ref{eqn: final_objective_4}).
The first term maximizes the chance of selecting robust grasps to execute.
The second and third terms are modified from~(\ref{eqn: final_objective_3}) to minimize the mean square error (MSE) between predictions and labels, as~\cite{cyclegan} suggests.
Such modifications stabilize the training process without loss of accuracy.

From all above, the grasp model $f_\theta (\cdot)$ is trained as:
\begin{equation} 
\begin{aligned} 
\label{eqn: final_loss}
    \theta = \underset{\theta}{\mathrm{argmin}}\ & \mathcal{L}(\hat{G}, \tilde{g})
\end{aligned} 
\end{equation}
where
\begin{equation*} 
\begin{aligned} 
    \hat{G} & = (\hat{Q}, \hat{\Phi}, \hat{W}) = f_\theta(I) \\
    \mathcal{L}(\hat{G}, \tilde{g}) & = \sum_{i=1}^k -\log P(\tilde{p_i} | \hat{Q}) +  \text{MSE}(\hat{\phi_i}, \tilde{\phi_i}) \\
    & \ \ \ \ \ \ \ \  +  \text{MSE}(\hat{w_i}, \tilde{w_i})
\end{aligned} 
\end{equation*}

The maximum likelihood grasp sampling loss (MLGSL) in (\ref{eqn: final_loss}) minimizes a pixel selection loss and two pixel-wise regression losses.
The main difference between MLGSL and others is MLGSL only uses existing labels.
Previous works~\cite{ggcnn,ggcnn2,andy_zeng_grasp,fcgqcnn,andy_grasp_push} regard unlabeled pixels as negatives and use regression or spatial cross-entropy losses to estimate the quality for the entire image.
In contrast, MLGSL only computes losses for pixels that have successful grasp labels (i.e., only at ${\tilde{p_i}}$), and it directly predicts whether a pixel is the best grasp point.
Specifically, MLGSL estimates the likelihood that a pixel is the most robust grasp with the grasp selection model and maximizes the likelihood for successful grasp labels.
As suggested in~\cite{q_learning_fail, rl_book}, estimating grasp qualities for each pixel is challenging due to only scarce label exists and the false-negative problem.
By contrast, MLGSL directly optimizes the likelihood for each grasp and is more reliable and easier to converge~\cite{rl_book}.

\subsection{Model Architectures}
The grasp planning model is used to predict dense grasp configuration maps $\hat{G}$, consisting of $\hat{Q}, \hat{\Phi}_{c}, \hat{\Phi}_{s}, \hat{W}$.
Note that angle maps $\Phi$ is calculated by $\Phi=\frac{1}{2} \tan^{-1} \frac{\Phi_{s}}{\Phi_{c}}$.
The model uses a fully convolutional topology, similar to~\cite{ggcnn2}.
The architecture includes four downsampling layers, two dilated layers, and two upsampling layers. 
Downsampling layers use kernel size of $[11, 5, 5, 5]$ respectively, activated by ReLU and max-pooling.
Two dilated layers apply $[5, 5]$ kernels with dilation $[2, 4]$.
Upsampling layers employ transpose convolutional kernels with size 3 and striding 2.

SAM blocks are also added as described in~\cite{sam}.
SAM utilizes both max-pooling and average-pooling along the channel axis and forwards them to a convolution layer. Outputs are then integrated into input features.

\subsection{Dataset}
The network is trained with a single object dataset and a cluttered object dataset.

This paper directly adopt Jacquard~\cite{jacquard} as the \textit{single object dataset}.
The dataset contains more than 50k images of 11k objects and 1 million success grasp labels. 
We apply random rotation and zoom to images and resize them to $300 \times 300$.

The \textit{cluttered object dataset} is generated based on Jacquard. 
A few images are randomly selected from the single object dataset and fused into a cluttered sample.
Before fusing, each single object image is segmented, rotated, zoomed, and translated in the image plane.
Success grasp labels are then merged and pruned according to collision constraints.
Since data in Jacquard includes images from different viewpoints, such operation can reflect the geometry of cluttered scenes.

\section{Training Results}
\label{sec: training}

We trained a series of models to test the proposed approach.
The goals of the experiments are three-fold:
1) to demonstrate that the proposed loss function can increase the grasp performance with fewer labels and samples,
2) to determine whether attention modules help in learning dense grasp configurations,
and 3) to inspect the collision-avoidance ability in cluttered scenes.

\subsection{Evaluation Metrics}
\label{subsec: eval_metric}
Three metrics are utilized to evaluate models' performance: predictions' success rate and predictions' accuracy and recall.
For prediction success rate, a predicted grasp $\hat{g_i}$ is considered success if
\begin{equation*} 
\begin{aligned}
\exists \tilde{g_j} \in \tilde{g}, \  \ni \text{IoU}(\tilde{g_j}, \hat{g_i}) \geq 25\% \text{   and } | \tilde{\phi_j}, \hat{\phi_i} | \leq 30\degree
\end{aligned}
\end{equation*}
where $\text{IoU}(\cdot)$ represents the intersection over union ratio between two grasps.
$\tilde{g}$ is the set of success grasp labels.
This paper selects top one (Top-1) and top five (Top-5) grasps to measure the success rate as in~\cite{ggcnn,ggcnn2,jacquard}.

Besides the grasp success rate, we measure predictions' accuracy and recall.
A grasp quality discriminator~\cite{dexnet2} was pre-trained to evaluate the robustness of grasps.
For each validation data, $100$ predicted grasps $\hat{g_i}$ are first uniformly sampled with predicted quality $\hat{q_i}$.
Then, the discriminator evaluates quality for $\hat{g_i}$, obtaining ground-truth quality label $q_i$.
Prediction accuracy and recall are measured based on $q_i$ and $\hat{q_i}$ for $i \in [1,...,100]$.

\subsection{Baseline Methods}
We compare the training performance of MLGSL to the following baseline approaches:

\subsubsection{Image-wise MSE (ImgMSE)} 
ImgMSE is introduced in~\cite{ggcnn} that use the same prediction maps $\hat{G}$ as ours.
This baseline augments successful grasp labels to $\bar{G}$ as in Section~\ref{subsec: dense_plan}.
The loss used to train the model is
\begin{equation*}
	\mathcal{L}_{\text{ImgMSE}} = \text{MSE}(\hat{Q}, \bar{Q}) + \text{MSE}(\hat{\Phi}, \bar{\Phi}) + \text{MSE}(\hat{W}, \bar{W})
\end{equation*}

\subsubsection{Maximum Likelihood Sampling with LogMSE (MLGSL+Log)} 
MLGSL+Log can be derived from~(\ref{eqn: final_objective_3}), which uses $\text{MSE}(\cdot)$ to replace $P(\cdot)$, i.e.
\begin{equation*} 
\begin{aligned} 
    \mathcal{L}_{\text{MLGSL+Log}} = \sum_{i=1}^k & - \log P(\tilde{p_i} | \hat{Q}) + \log \text{MSE}(\hat{\phi_i}, \tilde{\phi_i}) \\
    & +  \log \text{MSE}(\hat{w_i}, \tilde{w_i})
\end{aligned} 
\end{equation*}

\subsubsection{Pixel-wise MSE (PixMSE)} 
PixMSE applies supervisions solely on labeled pixels with MSE loss, i.e.
\begin{equation*} 
    \mathcal{L}_{\text{PixMSE}} = \sum_{i=1}^k \text{MSE}(\hat{q_i}, \tilde{q_i}) + \text{MSE}(\hat{\phi_i}, \tilde{\phi_i}) + \text{MSE}(\hat{w_i}, \tilde{w_i})
\end{equation*}

\subsection{Results}
For comparisons, we train variant models with different loss function designs and architectures.
Each model is trained with different seeds for 50 epochs to select the best one.

\subsubsection{Baseline Comparisons}
\begin{figure}[tb]
\begin{center}
	\includegraphics[width=2.7in]{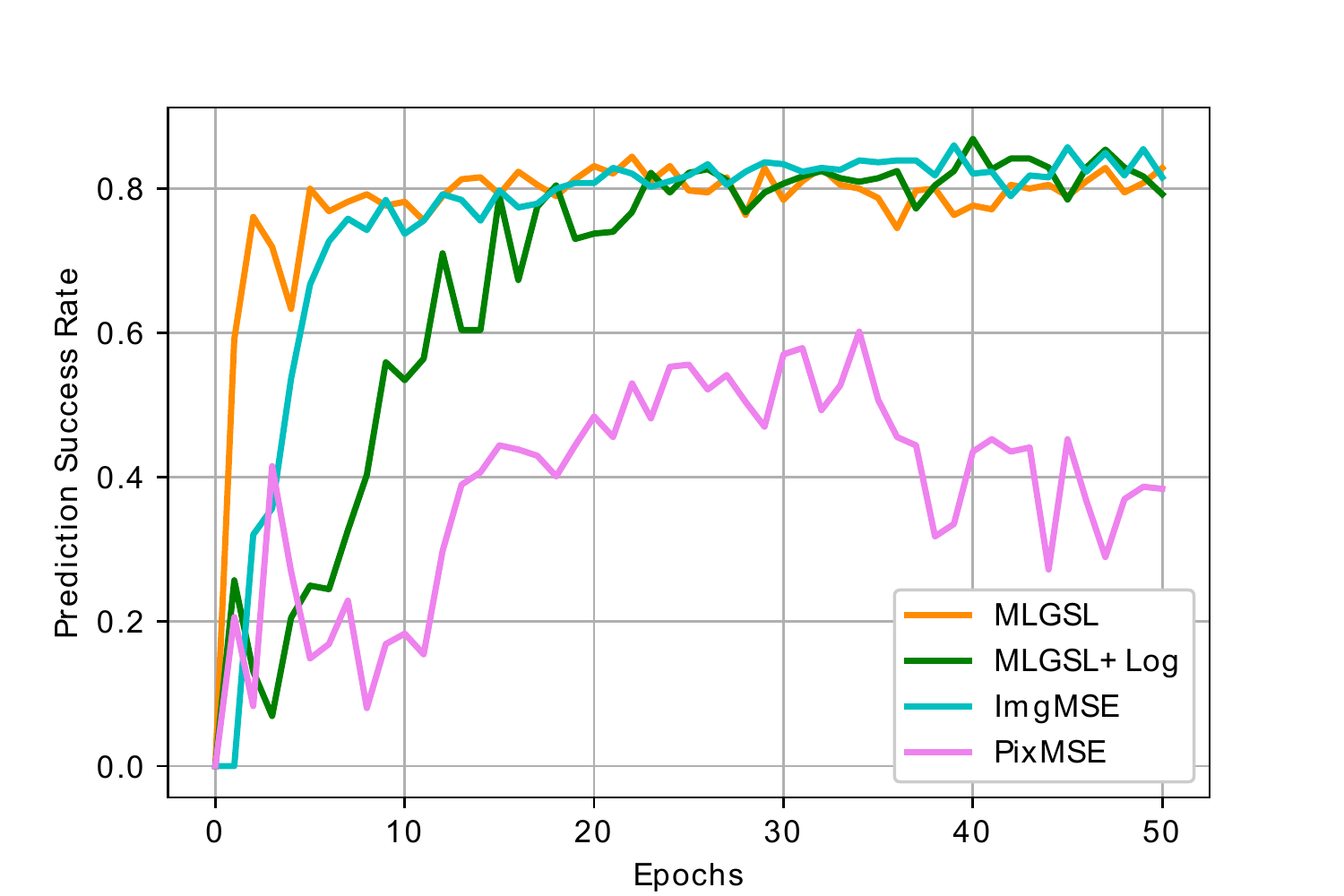}
	\vspace{-0.75em}
	\caption{Comparing the Top-1 prediction success rate of MLGSL with baseline methods.
	         Models are trained with densely-labeled datasets (16 labels per image).}
	\label{fig: baseline}
\end{center}
\end{figure}

\begin{table}[t]
\centering
\caption{Training performance of MLGSL and baselines (Mean \%)}
\label{tab: accuracy_recall}
\begin{tabular}{c|c|c|c|c}
\hline
Method                          & Top-1 & Top-5 & Accuracy & Recall \\ \hline
MLGSL                           & 82.8  & 91.0  & 80.2     & 75.7   \\
MLGSL (2 labels)                & 81.6  & 90.3  & 77.3     & 73.4   \\
ImgMSE~\cite{ggcnn2}            & 82.5  & 89.7  & 85.2     & 92.3   \\ 
ImgMSE~\cite{ggcnn2} (2 labels) & 42.4  & 45.2  & 41.9     & 17.4   \\ \hline
\end{tabular}
\end{table}

\begin{figure*}[t]
	\begin{center}
		\includegraphics[width=6.15in]{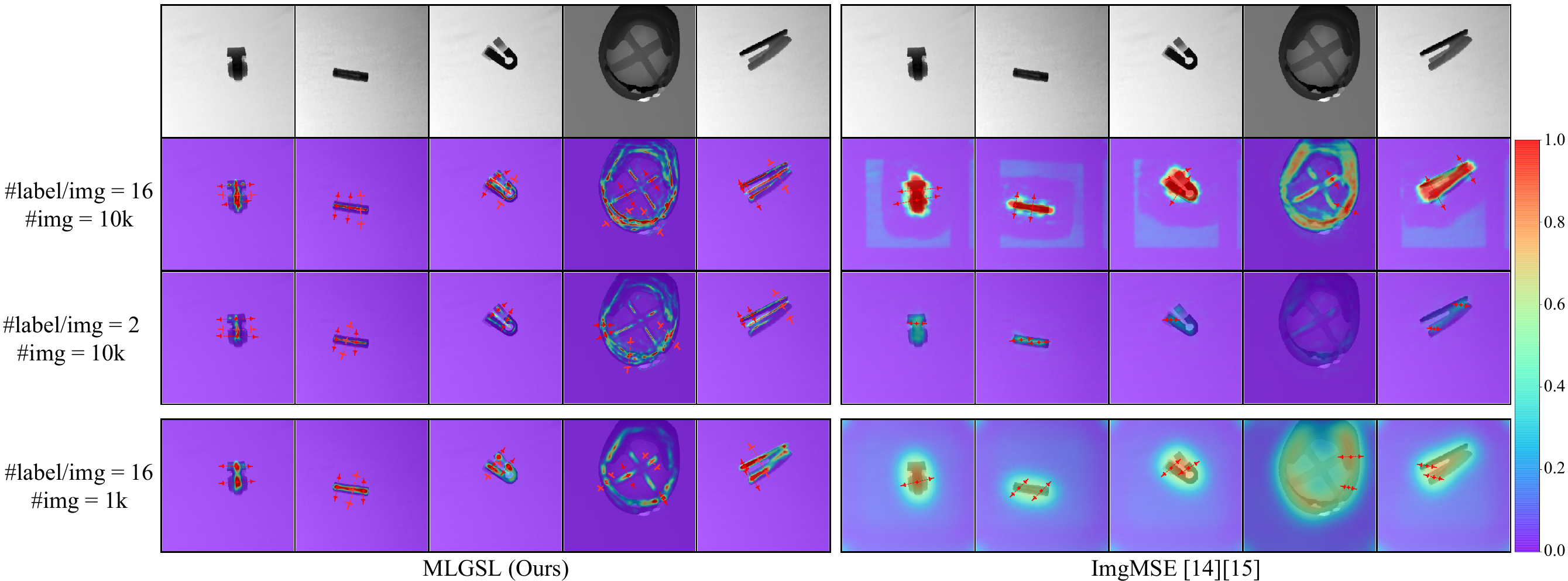}
		\vspace{-0.75em}
		\caption{Predicted grasp distributions with variant models. 
		        Predicted grasp qualities are painted as heatmaps with color listed in the right-sidebar.
                Detected grasps are labeled with red lines in each image.
		        (First row) input depth images, 
                (Left) results from models trained with MLGSL,
                (Right) results from models trained with ImgMSE,
                (Second row) results from datasets consisting of 16 labels per image and 10k training images,
                (Third row) results from datasets consisting of 2 labels per image and 10k training images,
                (Bottom) results from datasets consisting of 16 labels per image and 1k training images.
                }
		\label{fig: pred_g}
	\end{center}
\end{figure*}

Our first experiment compares MLGSL to three baseline methods with a single object dataset, in which each training sample includes 16 success grasp labels.
Top-1 and Top-5 prediction success rates are shown in Fig.~\ref{fig: baseline} and Table~\ref{tab: accuracy_recall}.
We see that MLGSL has similar performances compared to previous ImgMSE, while MLGSL shows a higher convergency rate at first a few epochs.
MLGSL with logarithm converges to a similar point as the previous two methods but with a slower rate, which might occur because the $\log(\cdot)$ operation lowers the gradient in each training step.
PixMSE performs the worst among the four approaches.
This likely due to it only applies supervisions on specific pixels, resulting in unbounded other areas.

We also compare MLGSL to ImgMSE on prediction accuracy and recall.
Results are shown in Table~\ref{tab: accuracy_recall}.
It is interesting to observe that models trained with MLGSL have lower accuracy and recall.
To seek reasons for such phenomenon, we plot several predicted $\hat{G}$ in the second rows of Fig.~\ref{fig: pred_g}.
As can be seen, models trained with MLGSL have a conservative estimation of the grasp quality and prefer to grasp each object's center.
This behavior leads to that only a small area has high quality (red area in Fig.~\ref{fig: pred_g}).
When measuring accuracy and recall, grasps are uniformly sampled in the image (Section~\ref{subsec: eval_metric}), making many of them out of the high-quality zone.
This fact then produces false-negative predictions and low accuracy and recall.

\subsubsection{Less Training Labels per Sample}
\begin{figure}[tb]
\begin{center}
	\includegraphics[width=2.7in]{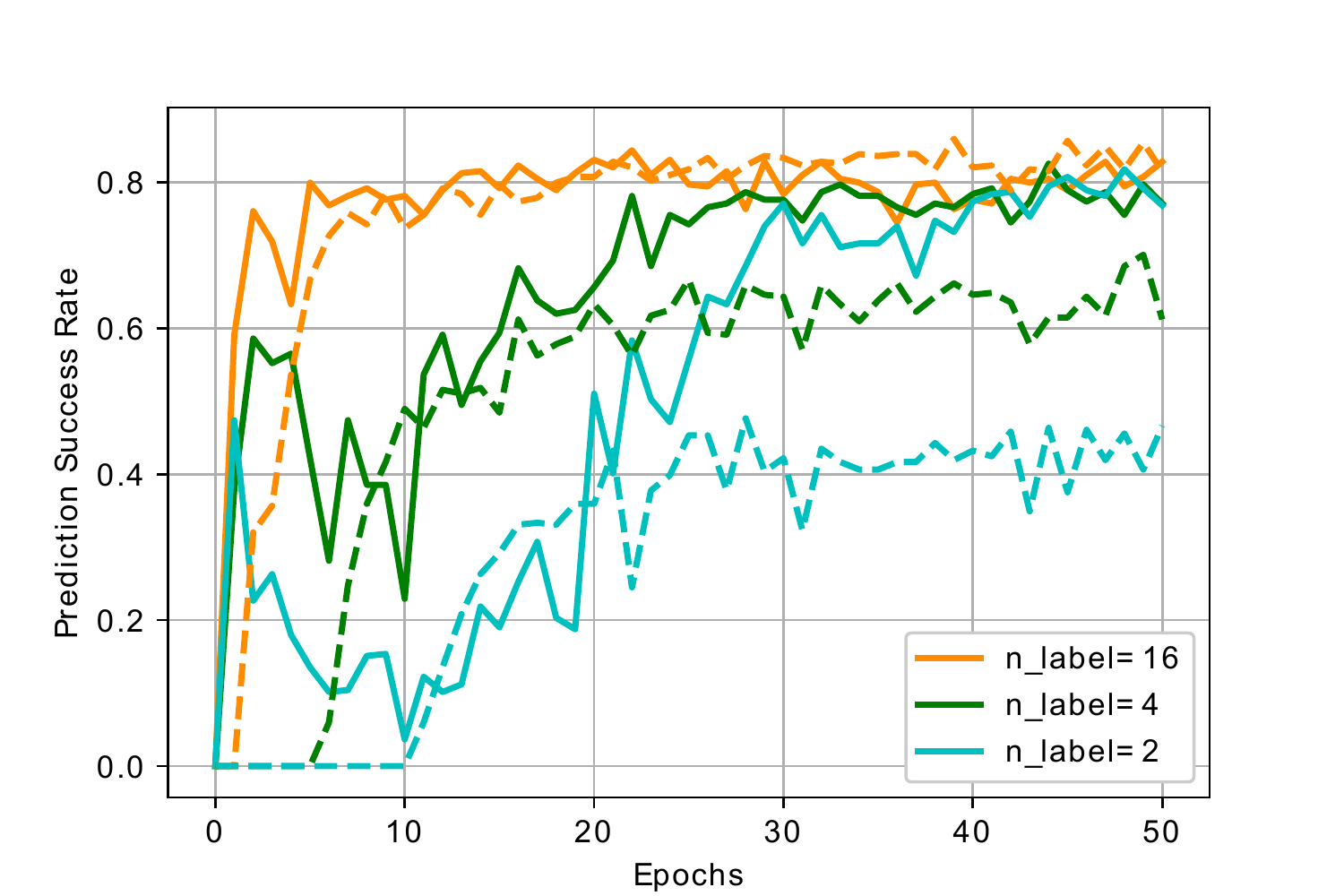}
	\vspace{-0.75em}
	\caption{Comparing the Top-1 prediction success rate of MLGSL to ImgMSE with different numbers of labels (n\_label).
Success grasp labels are down-sampled to $[2, 4]$ for each training image.
Solid lines indicate models' performance trained with MLGSL, and dashed lines indicate that trained with ImgMSE.}
	\label{fig: n_label}
\end{center}
\end{figure}

We then investigate whether our method can learn grasping with fewer labels.
For this study, we down-sample success grasp labels to $[2, 4]$ in each training data and still use all labels for validation.
It is a more difficult setting; the grasp planning model learns to effect change only through inadequate demonstrations.
We report results in Fig.~\ref{fig: n_label} and Table~\ref{tab: accuracy_recall}.
In the figure, models trained with MLGSL are evaluated with the Top-1 prediction success rate, indicated by solid lines.
Dashed lines indicate performances of models trained with ImgMSE.

From these results, we see that MLGSL is capable of learning to grasp with 2 labels per image, achieving prediction success rates at $81.6\%$ for Top-1 and $90.3\%$ for Top-5, which is similar to models trained with $16$ labels.
The third row in Fig.~\ref{fig: pred_g} shows predicted $\hat{G}$ by models trained with 2 labels per image.
We also notice that ImgMSE under-performs MLGSL in such settings.
This is attributed that the label generation method used by ImgMSE generate false-negatives.
Padded $\bar{G}$ may mistakenly label high-quality grasps to negatives since they are not close to existing success labels.
For MLGSL, unlabeled pixels are regulated indirectly with the probabilistic objective.
This procedure minimizes assumptions toward unlabeled areas, thus does not suffer from false-negatives.

\subsubsection{Less Training Samples}


We next train models with a smaller dataset (1k data) using MLGSL and ImgMSE.
The former makes models converge to $69.8\%$, while the latter converges at $60.2\%$.
The results suggest that less training data makes it harder to learn grasping strategies for both methods.
However, we still observe our MLGSL outperforms ImgMSE by about $10\%$.
We visualize prediction results at the bottom in Fig.~\ref{fig: pred_g}.
Compare to ImgMSE, MLGSL predicts more reasonable grasp distributions with less training data.
The less accurate predictions from ImgMSE might due to false training labels.
Models may require more data to compensate for wrong labels; thus, we observe better prediction results in previous experiments with more training data.

\subsubsection{Attention Module Integration}


Besides loss designs, we compare the effectiveness of attention modules by adding SAM to downsampling layers, upsampling layers, all layers, and no layers.
Interestingly, results suggest that the attention module is not a contributor to the dense grasp planning. 
All architectures converge to $81.6 \pm 1.2 \%$.
This could be because the backbone FCN is simple, which do not allow SAMs to take effect.
Furthermore, as suggested in~\cite{hsa, hsa2}, attentions might have a similar effect to quality map $\hat{Q}$ in our models.

\subsubsection{Collision-Free Datasets}


Networks in~\cite{ggcnn, ggcnn2} are trained with single object datasets and directly deployed to the clutter.
However, we observe collisions during grasp execution in practice.
To improve this, we train and test models with collision-free cluttered datasets.
The results show that the collision-free ratio improves to $84.2\%$ when training with the collision-free dataset compared to $67.3\%$ with the single object dataset.
Although we observe a $5\%$ down on prediction success rate, models trained with proposed datasets improve collision detection ability by $25\%$.

\section{Real-World Experiments}
\label{sec: exp}

\begin{figure}[tb]
\begin{center}
	\includegraphics[width=2.5in]{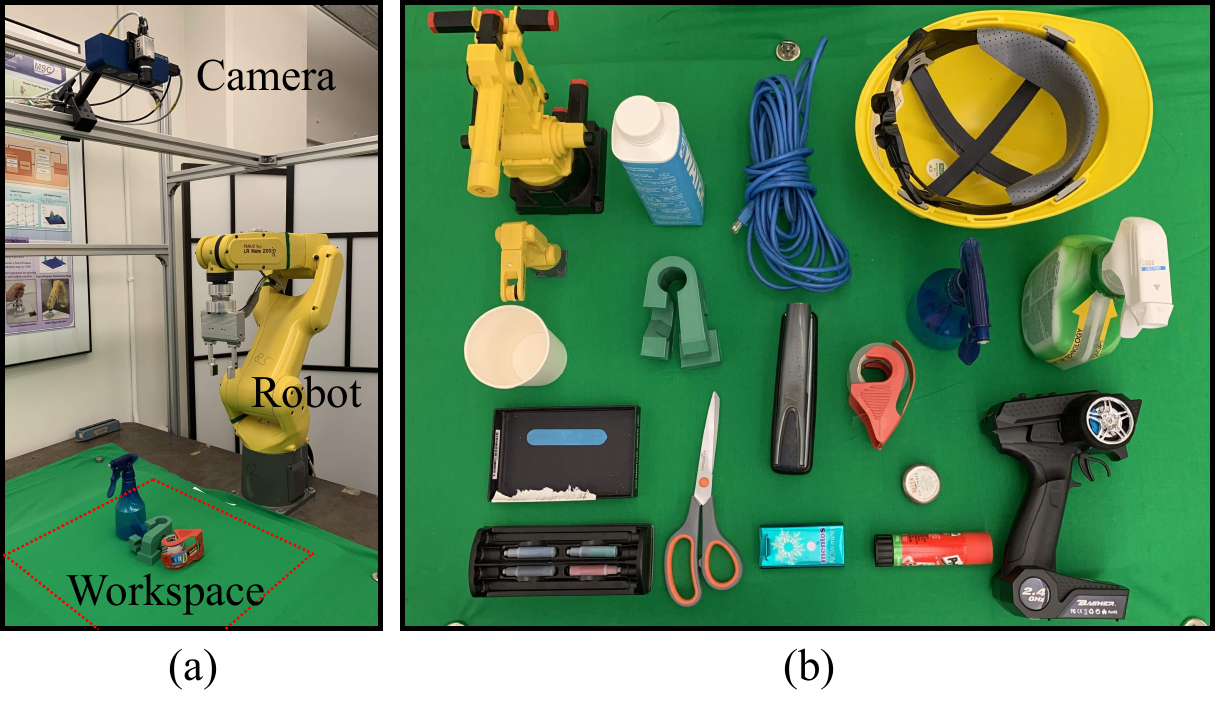}
	\vspace{-0.75em}
	\caption{(a) shows experimental setups.
	(b) shows objects used for single/cluttered grasping experiments.}
	\label{fig: exp_setup}
\end{center}
\end{figure}

Trained models were run on a laptop with GTX1060 GPU and 2.5GHz CPU. 
Experimental setups are shown in Fig.~\ref{fig: exp_setup}(a).
An Ensenso N35 camera was used to capture depth images.
Invalid depth values were inpainted using OpenCV~\cite{opencv}.
Robots executed the grasp with an offset $\epsilon=1$cm from the selected grasp point along the camera's z-axis.

17 household and 1 adversarial objects were selected to test the models (Fig.~\ref{fig: exp_setup}(b)).
Household objects contain items of varying sizes and shapes.
Most of the items (staple, tape, cube, robots, sprayers, glue stick) appear in previous works.
We used several additional objects that are deformable (cable) and perceptually challenging (thin edges on the cup, helmet, board eraser, scissor, and reflective zinc container).
We also added an adversarial object~\cite{dexnet2} to verify models' robustness with a complex geometry.

\begin{table}[t]
\centering
\caption{Real-World performance of MLGSL and Baselines}
\label{tab: real_results}
\begin{tabular}{c|c|c|c}
\hline
Method                       & Required Data                 & SR (\%)       & CTPG (ms)              \\ \hline
GGCNN~\cite{ggcnn}           & 10k $\times$ 16               & 83.3 (30/36)  & 21      \\
GGCNN2~\cite{ggcnn2}         & 10k $\times$ 16               & 88.9 (48/54)  & 21      \\
GQCNN~\cite{dexnet2}         & 7m $\times$ 1                 & 91.7 (33/36)  & 570      \\
FC-GQCNN~\cite{fcgqcnn}      & 7m $\times$ 1                 & 90.7 (49/54)  & 29      \\
CGPN~\cite{cgpn}             & 30k $\times$ 20               & 85.2 (46/54)  & 410      \\
MLGSL (Ours)                 & 10k $\times$ 2                & 90.7 (98/108)  & 21      \\ \hline
\end{tabular}
\vspace{0.5em}

Require data represents the size of training datasets (number of images $\times$ number of labels per image); SR and CTPG represent grasp success rate and computation time per grasp respectively.
\end{table}

We performed physical grasping experiments with two arrangements: 1) isolated single objects and 2) cluttered objects.
For single object grasping, we trained a model with MLGSL, and a dataset consists of 2 labels per image.
We compared our approach with five baselines~\cite{ggcnn, ggcnn2, dexnet2, fcgqcnn, cgpn}.
Results are reported in Table~\ref{tab: real_results}. Each object was grasped for the same time.
Note that all baselines require larger well-labeled datasets to achieve listed performances.

For cluttered scenarios, objects were randomly placed inside the workspace.
The robot attempted one grasp each time, and the grasped object was removed from the scene.
The picking order is greedily determined to maximize the grasp success rate of overall objects and avoid collisions.
This procedure continues until all objects are removed or consecutively failed five times.
We ran this experiment 10 times to measure performances.
Models trained with MLGSL and cluttered datasets achieved an object removal rate of 90\%, compared to 70\% in models trained with single datasets, mainly due to undetected collisions.
Comparing to~\cite{ggcnn2}, we observed similar results that models trained with cluttered datasets outperform that with single datasets.

\begin{figure}[tb]
\begin{center}
	\includegraphics[width=2.8in]{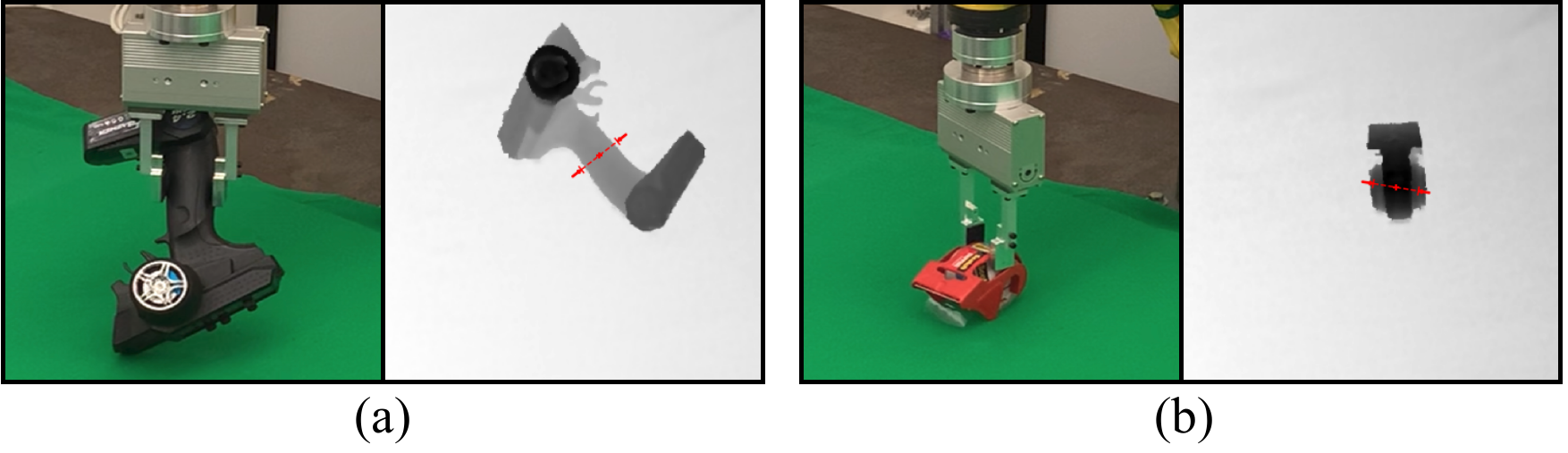}
	\vspace{-0.75em}
	\caption{Two failure cases.
	(a) shows the object slippage when the robot grasps heavy objects, 
	and (b) shows models mistakenly generate grasps toward deformable thin covers.}
	\label{fig: failures}
\end{center}
\end{figure}

Figure~\ref{fig: failures} displays two common failures of MLGSL.
One failure mode occurs when the object is heavy.
The current method assumes a fixed contact force, and heavy objects can slip without a grasp force controller.
The second type of failure occurs when a thin deformable layer is on top of the object's main body.
It is challenging to distinguish thin layers from solid cubes in depth images.
Such ambiguity tricks the model into generating unstable grasps.

\section{Discussion and Conclusion} 
\label{sec: conclusion}
This paper tackles the data sparsity issue in grasp planning with several key contributions.
First, we propose a stochastic process to select grasps with dense grasp planners.
Second, we present the MLGSL to train FCNs with a small empirical dataset; we show that it can match the performance of the state-of-the-art methods with fewer training data.
Third, we show that the attention mechanisms are not contributing to the dense grasp planning.
Lastly, we provide a grasping dataset to improve the performance of networks in clutter.

The present work also has limitations.
This paper only shows experiments on low-DoF grasping tasks.
However, the proposed MLGSL does not constrain grasp dimensions and thus can be generalized to high-DoF grasps.
Validating MLGSL in high-DoF tasks will be our future works.
Moreover, collision-free is not guaranteed with the proposed dataset; a collision-pruning module is still required in practice.



\addtolength{\textheight}{-1cm}   



\newpage
\bibliographystyle{IEEEtran}
\bibliography{references}

\end{document}